\documentclass[review]{elsarticle}

\usepackage{lineno,hyperref}
\usepackage{color}
\usepackage{array}
\usepackage{booktabs}
\usepackage{threeparttable}
\usepackage[T1]{fontenc}
\usepackage{graphicx}

\newcommand{\PreserveBackslash}[1]{\let\temp=\\#1\let\\=\temp}
\modulolinenumbers[5]

\journal{International Journal of Critical Infrastructure Protection}

\begin{document}

\begin{frontmatter}

\title{Modeling contaminant intrusion in water distribution networks based on D numbers }


\author{\footnotesize Li Gou$^a$, Yong Deng$^{a,b,}$\footnote{Corresponding author: Yong Deng, School of Computer and Information Science, Southwest University, Chongqing, 400715, China. Email address: ydeng@swu.edu.cn; prof.deng@hotmail.com}, Rehan Sadiq$^c$, Sankaran Mahadevan$^b$, }

\address{$^a$School of Computer and Information Science,\\
 Southwest University, Chongqing 400715, China\\
$^b$School of Engineering, Vanderbilt University, Nashville, TN 37235, USA\\
$^c$School of Engineering, University of British Columbia Okanagan,\\
 3333 University Way, Kelowna, BC, Canada V1V 1V7}
\begin{abstract}
Efficient modeling on uncertain information plays an important role in estimating the risk of contaminant intrusion in water distribution networks. Dempster-Shafer evidence theory is one of the most commonly used methods. However, the Dempster-Shafer evidence theory has some hypotheses including the exclusive property of the elements in the frame of discernment, which may not be consistent with the real world. In this paper, based on a more effective representation of uncertainty, called D numbers, a new method that allows the elements in the frame of discernment to be non-exclusive is proposed. To demonstrate the efficiency of the proposed method, we apply it to the water distribution networks to estimate the risk of contaminant intrusion.
\end{abstract}

\begin{keyword}
contaminant intrusion\sep water distribution networks\sep Dempster-Shafer evidence theory\sep D numbets\sep fuzzy numbers\sep belief function
\end{keyword}

\end{frontmatter}

\linenumbers

\section{Introduction}
Water supply systems are one of the most important fundamentals for human living and development\cite{sargaonkar2013model,nyende2013application,vairavamoorthy2007modelling}. The topic relating to the performance of the water supply systems under varied conditions has been paid considerable attention \cite{preis2008multiobjective,tamminen2008water,perelman2009extreme,islam2013evaluating,el2004fuzzy}. Water supply systems are usually designed, constructed, operated, and managed in an open environment. As a result, they are inevitably exposed to varied uncertain threats and hazards \cite{walski2003advanced,islam2013evaluation,xin2012hazard,khanal2006distribution}. Contaminant intrusion in a water distribution network is a complex phenomenon, which depends on three elements - a pathway, a driving force and a contamination source \cite{lindley2001framework,lindley2002assessing,rasekh2014drinking,shen2011false,preis2007contamination}. However, the data on these elements are generally incomplete, non-specific and uncertain \cite{sadiq2008predicting}.

Quantitative aggregation of incomplete, uncertain and imprecise information data warrants the use of soft computing methods \cite{zadeh1984review,sadiq2006estimating,jenelius2010critical}. Soft computing methods such as fuzzy set theory \cite{zadeh1965fuzzy,deng2012fuzzy,zhang2013ifsjsp,aghaarabi2014compara,setola2009critical,oliva2011fuzzy}, rough set \cite{pawlak2007rudiments,pawlak2007rough}, Dempster-Shafer evidence theory \cite{dempster1967upper,shafer1976mathematical,wei2013identifying} can essentially provide rational solutions for complex real-world problems. The traditional Bayesian (subjectivist) probability approach cannot differentiate between aleatory and epistemic uncertainties and is unable to handle non-specific, ambiguous and conflicting information without making strong assumptions. These limitations can be addressed by the application of Dempster-Shafer evidence theory, which was found to be flexible enough to combine the rigor of probability theory with the flexibility of rule-based systems \cite{sadiq2006estimating,deng2014environmental,huang2013new}.

Due to the requirements of safety and reliability of the water supply system, risk assessment has been recognised as a useful tool to identify threats, analyse vulnerabilities and risks, and select mitigation measures for water supply systems \cite{li2007hierarchical,weickgenannt2010risk}. Accordingly, an object-oriented approach for water supply systems is proposed in \cite{li2007hierarchical}, which is based on aggregative risk assessment (similar to \cite{sadiq2004aggregative}) and fuzzy fault tree analysis and use fuzzy evidential reasoning method to determine the risk levels associated with components, subsystems, and the overall water supply system. Then evidential reasoning model based on Dempster-Shafer evidence theory is applied to estimate risk of contaminant intrusion in water distribution network \cite{sadiq2006estimating,deng2011modeling}.

However, there are some shortcomings in previous methods. In the classical Dempster-Shafer evidence theory, a problem domain is indicated by the concept of frame of discernment, and the concept of basic probability assignment (BPA) is used to represent the uncertain information. But there are several hard hypotheses and constraints. For example, the elements in the frame of discernment must be mutually exclusive and the sum of BPA must be equal to 1, which is usually inconsistent with the applications above. These shortcomings have greatly limited its practical application \cite{deng2014environmental,deng2012d}.

Recently, a new methodology called D numbers \cite{deng2014environmental,deng2012d,Deng2014DAHPSupplier,Deng2014BridgeDNs} to represent uncertain information has been proposed, which is an extension of Dempster-Shafer evidence theory. D numbers can effectively represent uncertain information. The exclusive property of the elements in the frame of discernment is not required, and the completeness constraint is released. Due to the propositions of applications in the real word could not be strictly mutually exclusive, these two improvements are greatly beneficial. To get a more accurate uncertain data fusion, a discounting of D numbers based on the exclusive degree is necessary. In this paper, a new exclusive model based on D numbers to combine the uncertain data at a non-exclusive level was proposed.

The rest of the paper is organized as follows. Section 2, some definitions of D numbers is introduced. A step-by-step applications of the proposed model to a numerical example are illustrated in Section 3. In Section 4, the proposal exclusive model based on D numbers was applied to the water distribution networks to assess the risk of contaminant intrusion. Conclusions are given in Section 5.

\section{D numbers}\label{D numbers}
Situations in the real world are affected by many sources of uncertainty. Many existing theories have been developed to model various types of uncertainty with some desirable properties. However, these theories still contain deficiencies that can not be ignored. For example, due to the inherent advantages in the representation and handling of uncertain information, the Dempster-Shafer evidence theory is being studied for use in many fields. Such as decision making, pattern recognition\cite{tao2012identification}, risk assessment \cite{sadiq2006estimating,deng2011risk}, supplier selection and others. In the mathematical framework of Dempster-Shafer theory, the basic probability assignment (BPA) defined on the frame of discernment is used to express the uncertainty quantitatively. A problem domain indicated by a finite and mutually exclusive non-empty set $\Omega$ is called a frame of discernment. Let $2^\Omega$ denote the power set of $\Omega$. The elements in the $2^\Omega$ are called propositions. The BPA is a mapping $m$ from $2^\Omega$ to $[0,1]$, and satisfying the following condition \cite{dempster1967upper,shafer1976mathematical}:
\begin{equation}\label{mass function}
  m(\phi)=0 \quad  and \quad  \sum_{A\in{2^\Omega}}{m(A)=1}
\end{equation}

\begin{figure}[htbp]
\begin{center}
\centerline{
\includegraphics[width=8cm]{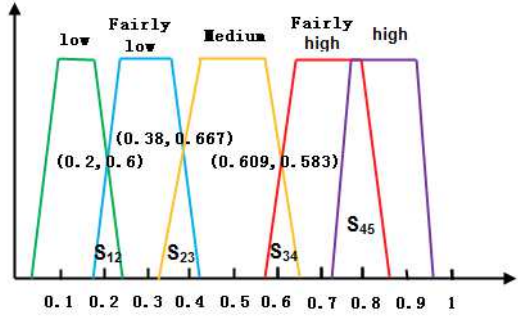}
}
\caption{Five-granular set of linguistic constants.
}\label{fuzzy_figure}
\vspace{0.2in}
\end{center}
\end{figure}

BPA has an advantage of directly expressing the `uncertainty' by assigning the basic probability to the subsets of the set composed of $N$ individual objects, rather than to each of the individual objects. But there exists some strong hypotheses and hard constraints on the frame of discernment and BPA, which limit the practical application of Dempster-Shafer evidence theory. One of the hypotheses is that the elements in the frame of discernment are required to be mutually exclusive. However, this hypothesis is difficult to be satisfied in many situations. For example, the linguistic assessments shown in Fig. \ref{fuzzy_figure} can be "Low", "Fairly low", "Medium", "Fairly high", "High". Due to these assessments is based on human judgment, they inevitably contain intersections. The exclusiveness between these propositions can't be guaranteed precisely, so that the application of Dempster-Shafer evidence theory is questionable and limited in this situation. That means, it is not correct to give a BPA like this: $m({Fairly\quad high},{High})=0.8,\quad m({High})=0.2$.

D numbers\cite{deng2012d} is a new representation of uncertain information, which is an extension of Dempster-Shafer evidence theory. It overcomes the existing deficiencies in Dempster-Shafer evidence theory and appears to be more effective in representing various types of uncertainty. D numbers are defined as follows.

Let $\Omega$ be a finite nonempty set, a D number is a mapping formulated by
\begin{equation}\label{D mapping}
  D:\Omega\rightarrow[0,1]
\end{equation}
with
\begin{equation}\label{function}
  \sum_{B\subseteq\Omega}{D(B)\leq1} \quad  and \quad D(\phi)=0
\end{equation}

It seems that the definition of D numbers is similar to the definition of BPA. But note that, the first difference is the concept of the frame of discernment in Dempster-Shafer evidence theory. The elements in the frame of discernment $\Omega$ of D numbers do not require mutually exclusive. Second, the completeness constraint is released in D numbers. If $\sum_{B\subseteq\Omega}{D(B)=1}$, the information is said to be complete; if $\sum_{B\subseteq\Omega}{D(B)\leq1}$, the information is said to be incomplete. An illustrative example is given to show the D numbers as below.

\textbf{Example 1.} Suppose a project is assessed, the assessment score is represented by an interval [0, 100]. In the frame of Dempster-Shafer evidence theory, an expert could give a BPA to express his assessment result:
$$m(\{a_1\})=0.2$$
$$m(\{a_3\})=0.7$$
$$m(\{a_1,a_2,a_3\})=0.1$$
where $a_1=[0,40]$, $a_2=[41,70]$, $a_3=[71,100]$. The set of $\{a_1,a_2,a_3\}$ is a frame of discernment in Dempster-Shafer evidence theory.

However, if another expert gives his assessment result by using D numbers, it could be:
$$D(\{b_1\})=0.2$$
$$D(\{b_3\})=0.6$$
$$D(\{b_1,b_2,b_3\})=0.1$$
where $b_1=[0,45]$, $b_2=[38,73]$, $b_3=[61,100]$. Note that the probability assignment of the set of $\{b_1,b_2,b_3\}$ is not a BPA, because the elements in the set $\{b_1,b_2,b_3\}$ are not mutually exclusive. Due to $D(\{b_1\}) +D(\{b_3\}) +D(\{b_1,b_2,b_3\}) = 0.9$, the information is incomplete. This example has shown the differences between BPA and D numbers.

If a problem domain is $\Omega=\{b_1,b_2,...,b_i,...,b_n\}$, where $b_i\in{R}$ and $b_i\neq{b_j}$ if $i\neq{j}$, a special form of D numbers can be expressed by:
$$D(\{b_1\})=v_1$$
$$D(\{b_2\})=v_2$$
$$...$$
$$D(\{b_i\})=v_i$$
$$...$$
$$D(\{b_n\})=v_n$$
or simply denoted as $D=\{(b_1,v_1), (b_2,v_2),...,(b_i,v_i),...,(b_n,v_n)\}$, where $v_i > 0$ and $\sum_{i=1}^n{v_i\leq1}$. Some properties of D numbers are introduced as follows.

\begin{figure}[htbp]
\begin{center}
\centerline{
\includegraphics[width=10cm]{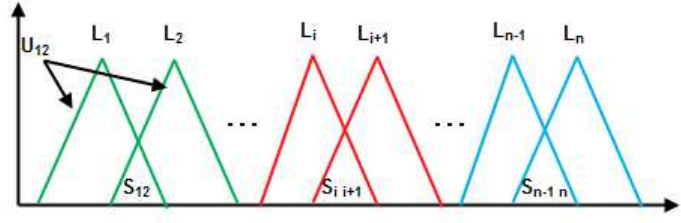}
}
\caption{Four linguistic constants.
}\label{linguistic example}
\vspace{0.2in}
\end{center}
\end{figure}

\textbf{Relative matrix.} $n$ linguistic constants expressed in normal triangular fuzzy numbers are illustrated in Fig. \ref{linguistic example}. The area of intersection $S_{ij}$ and union $U_{ij}$  between any two triangular fuzzy numbers $L_i$ and $L_j$ can be can calculated to represent the non-exclusive degree between two D numbers. For example, the intersection $S_{12}$ and the union $U_{12}$ in Fig. \ref{linguistic example}. The non-exclusive degree $D_{ij}$ can be calculated as follows:
\begin{equation}\label{exclusive degree}
  D_{ij}=\frac{S_{ij}}{U_{ij}}
\end{equation}

 It should be emphasized that how to determine the non-exclusive degree depends on the application type. Due to the characteristic of the fuzzy numbers, we choose the area of intersection and union between two fuzzy numbers. A relative Matrix for these elements based on the non-exclusive degree can be build as below:
\begin{equation}\label{relative matrix example}
  R=
\left[
\begin{array}{ccccccc}
      &  L_1     &  L_2    & \ldots  &  L_i    &\ldots & L_n\\
 L_1  & 1        &  D_{12} &\ldots  &  D_{1i} &\ldots  &  D_{1n}  \\
 L_2  &  D_{21}  &  1      &\ldots  &  D_{2i} &\ldots  &  D_{2n}  \\
 \vdots  & \vdots  & \vdots &\ldots      &\vdots     &\ldots&  \vdots  \\
 L_i  &  D_{i1}  &  D_{i2} &\ldots  &  1 &\ldots &  D_{in}\\
 \vdots  & \vdots  & \vdots &\ldots      &\vdots     &\ldots&  \vdots  \\
 L_n  &  D_{n1}  &  D_{n2} &\ldots  &  D_{ni} &\ldots & 1
\end{array}
\right]
\end{equation}


\textbf{Exclusive coefficient.} The exclusive coefficient $\varepsilon$ is used to characterize the exclusive degree of the propositions in a assessment situation, which is got by calculating the average non-exclusive degree of these elements using the upper triangular of the relative matrix. Namely:
\begin{equation}\label{ce}
  \varepsilon=\frac{\sum_{i,j=1,i\neq{j}}^{n}{D_{ij}}}{n(n-1)/2}
\end{equation}
where $n$ is the number of the propositions in the assessment situation. Smaller the $\varepsilon$ is, the more exclusive the propositions of the application are. When $\varepsilon=0$, the propositions of application are completely mutually exclusive. That is, this situation is up to the requirements of the Dempster-Shafer evidence theory.

\textbf{The combination rule of D numbers.}  Firstly, the given D numbers should be discounted by the exclusive coefficient $\varepsilon$, which can guarantee the elements in the frame of discernment $\Omega$ to be exclusive. The D numbers can be discounted as below:
      \begin{eqnarray*}
        D(A_i)_\varepsilon=D(A_i).(1-\varepsilon)
      \end{eqnarray*}
      \begin{equation}\label{discount}
        D(\Theta)_\varepsilon=D(\Theta).(1-\varepsilon)+\varepsilon
      \end{equation}
      where $A_i$ is the elements in $\Omega$.

Then the combination rule of D numbers based on the exclusive coefficient is illustrated as follows.
\begin{equation}\label{conbination}
  D(A)_{\varepsilon}=\frac{\sum_{B\cap{C}=A}{D_1(B)_{\varepsilon}D_2(C)_{\varepsilon}}}{1-k}
\end{equation}
with
\begin{equation}\label{conflict}
  k=\sum_{B\cap{C}={\phi}}{D_1(B)_{\varepsilon}D_2(C)_{\varepsilon}}
\end{equation}
where $k$ is a normalization constant, called conflict because it measures the degree of conflict between $D_1$ and $D_2$.

One should note that, if $\varepsilon=0$, i.e, the elements in the frame of discernment $\Omega$ are completely mutually exclusive, the D numbers will not be discount by the exclusive coefficient. That is, the mutually exclusive situation of D numbers is completely the same with the Dempster-Shafer evidence theory.

\section{Proposed model based on D numbers}\label{exclusive section}
In this section, the proposed model based on D number will be illustrated with a numerical example. In the model, the exclusive coefficient is proposed to represent the exclusive degree among the propositions in the frame of the discernment. From section \ref{D numbers}, we know that, one of the advantages of D numbers is that the elements in the frame of the discernment $\Omega$ are not required to be mutually exclusive. It's clear that propositions of application in real world can't be completely mutually exclusive, so define the exclusive coefficient to undermine the non-exclusive property is essential. For example, linguistic assessment based on human judgment can be "fairly good", "good" and "very good", which is obviously non-exclusive but need to be assessed. A numerical example to show the proposed method is illustrated through a step-by-step description.

\textbf{Step 1} Constructing the linguistic constants \cite{deng2011risk} expressed in positive trapezoidal fuzzy numbers. The details of linguistic constants presented in Fig. \ref{fuzzy_figure} are shown in Table \ref{linguistic table}.

\begin{table}[htbp]
\centering
\def\~{\hphantom{0}}
\caption{Linguistic constants represented by trapezoidal fuzzy numbers}
\label{linguistic table}
\begin{tabular*}{\columnwidth}{@{}l@{\extracolsep{\fill}}c@{\extracolsep{\fill}}r@{\extracolsep{\fill}}r@{\extracolsep{\fill}}r@{\extracolsep{\fill}}r@{\extracolsep{\fill}}l@{\extracolsep{\fill}}c@{\extracolsep{\fill}}c@{\extracolsep{\fill}}c@{}}
\hline
{Variable}
linguistic constants  & Fuzzy numbers\\
\hline
Low  & (0.04,0.1,0.18,0.23)\\
Fairly low  & (0.17,0.22,0.36,0.42)\\
Medium  & (0.32,0.41,0.58,0.65)\\
Fairly high  & (0.58,0.63,0.80,0.86)\\
High  & (0.72,0.78,0.92,0.97)\\
\hline
\end{tabular*}
\end{table}

\textbf{Step 2} Calculate the area of intersection $S_{i,j}$ and the union $U_{i,j}$ between any two trapezoidal fuzzy numbers $D_i$ and $D_j$ respectively. For example, $S_{12}, S_{23}, S_{34}$ and $S_{45}$ in the Fig. \ref{fuzzy_figure} is some intersections.

\textbf{Step 3} Calculate the non-exclusive degree $D_{ij}$ between two fuzzy numbers according to Eq. \ref{exclusive degree}. For example, the non-exclusive degree between "low" and "fairly low" can be calculated as follows:
      $$D_{12}=\frac{0.018}{0.312}=0.0577$$

\textbf{Step 4} Build a relative Matrix following the regulation defined in Section \ref{D numbers}. The relative Matrix of current example is:
  \begin{equation}\label{relative matrix}
  R=
\left[
\begin{array}{cccccc}
  & low & fairly\quad low & medium & firly\quad high & high\\
     low & 1 & 0.0577 & 0 & 0 & 0 \\
     fairly\quad low & 0.0577 & 1 & 0.081 & 0 & 0\\
     medium & 0 & 0.081 & 1 & 0.0449 & 0 \\
     firly\quad high & 0 & 0 & 0.0449 & 1 & 0.2353\\
     high & 0 & 0 & 0 & 0.2353 & 1
\end{array}
\right]
\end{equation}

  \textbf{Step 5} Calculate the exclusive coefficient $\varepsilon$ through Eq. \ref{ce}. That is:
  $$\varepsilon=\frac{0.0577+0.081+0.0449+0.2353}{10}=0.042$$

  \textbf{Step 6} Discounting the D numbers according to Eq. \ref{discount}. Two D number $D_1$ and $D_2$ based on the fuzzy numbers in Table \ref{linguistic table} are shown in Table \ref{discount D number}. Then we can discount these two D numbers using the exclusive coefficient $\varepsilon$, the results $D(A_i)_\varepsilon$ are also in Table \ref{discount D number}.

\begin{table}[htbp]
\centering
\def\~{\hphantom{0}}
\caption{The initial and discounted D numbers}
\label{discount D number}
\begin{tabular*}{\columnwidth}{@{}l@{\extracolsep{\fill}}c@{\extracolsep{\fill}}r@{\extracolsep{\fill}}r@{\extracolsep{\fill}}r@{\extracolsep{\fill}}r@{\extracolsep{\fill}}l@{\extracolsep{\fill}}c@{\extracolsep{\fill}}c@{\extracolsep{\fill}}c@{}}
\hline
$A_i$ & $D_1(A_i)$  & $D_2(A_i)$ & $D_1(A_i)_\varepsilon$  & $D_2(A_i)_\varepsilon$\\
\hline
 $(\{low\})$                   & 0.12  & 0.1 & 0.115 & 0.096\\
 $(\{low\},\{fairly\quad low)$ & 0.7   & 0.06 & 0.671 & 0.057\\
 $(\{medium\})$                & 0.02  & 0.6 &0.019 & 0.575\\
 $(\{high\},\{medium\})$       & 0.1   & 0.2 &0.096 & 0.192\\
 $(\{fairly\quad high\})$      & 0.06 & 0.04 &0.057 &0.38\\
 $\{\Theta\}$                  & 0 & 0 &0.042&0.042\\
\hline
\end{tabular*}
\end{table}

\textbf{Step 7} Use the combination rule of D numbers given in Section \ref{D numbers} to get the final assessment. Namely:
$$D(\{low\})=0.3096$$
$$D(\{low\},\{fairly\quad low\})=0.2359$$
$$ D(\{medium\})=0.3232$$
$$D(\{fairly\quad high\})=0.0213$$
$$D(\{high\},\{medium\})=0.1039$$
$$ D(\{\Theta\})=0.0061$$

\section{Estimating risk of contaminant intrusion}
Contaminant intrusion in a water distribution network is a complex phenomenon, which depends on three elements C a pathway, a driving force and a
contamination source \cite{lindley2001framework,lindley2002assessing}. However, the data on these elements are generally incomplete, non-specific and uncertain. In earlier studies, evidential reasoning model has been used to estimate risk of contaminant intrusion in water distribution network based on above three elements \cite{sadiq2006estimating,deng2011modeling}. This section provides another methods called D numbers to assess the risk of contaminant intrusion in distribution networks.

In previous work \cite{sadiq2006estimating,deng2011modeling}, the problem domain of risk of an intrusion can be described by a universal set $\Theta= \{P, NP\}$, in which `P' denotes `possible' and `NP' denotes `not possible' intrude. The power set of the risk of intrusion consists of two singletons $\{P\}$ and, $\{NP\}$, a universal set $\{P,NP\}$ and the empty set $\{\phi\}$. As described earlier, the risk of contaminant intrusion can be evaluated based on three elements ¨C a pathway ($e_1$), a driving force ($e_2$), and a contamination source ($e_3$).

We select surrogate measures to simplify the intrusion problem. The breakage rate (\# of breaks/100 km/year) is taken as a surrogate measure for an "intrusion pathway", transient pressure(psi) is taken as a surrogate for a "driving force", and the separation distance (meters) between a contaminant source and a water main as a surrogate measure for the "source of contamination". The frame of discernment $\Omega$ are mapped to obtain D numbers(i.e., $D_1, D_2, and D_3$), where each of them can be assigned to these subsets {P}, {P, NP}, and {NP}.

\begin{figure}[htbp]
\begin{center}
\centerline{
\includegraphics[width=12cm]{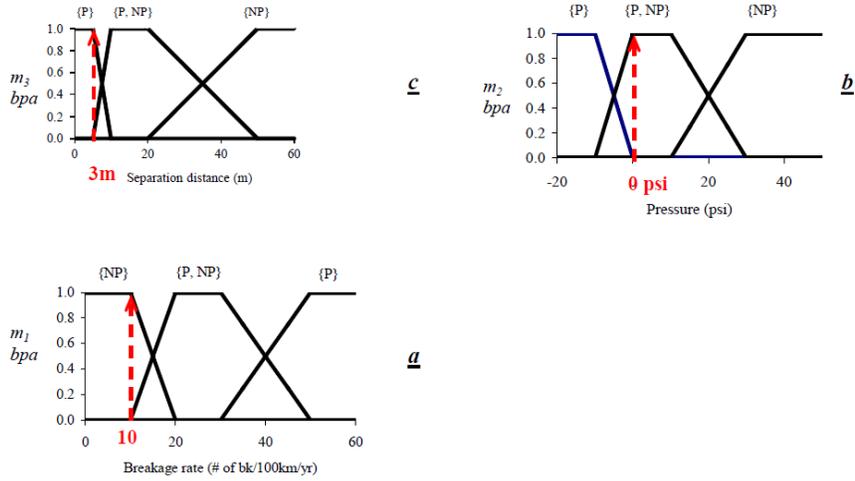}
}
\caption{Three evidence bodies: pathway, driving force and contamination source.
}\label{bpa_figure}
\vspace{0.2in}
\end{center}
\end{figure}

For this problem, we propose a new contaminant intrusion model based on D numbers, which make use of the exclusive coefficient to represent the exclusive degree among the propositions of the water distribute networks. A step-by-step description is provided below:

\textbf{Step 1} Construct the description of propositions and collect data. For example, in Fig. \ref{bpa_figure}, the description of each proposition is represented as trapezoidal fuzzy numbers. The collected data are real numbers. It should be emphasized here that triangular fuzzy numbers and real numbers are the special cases of generalized trapezoidal fuzzy numbers. This step is the same as in case of \cite{sadiq2006estimating}.

\textbf{Step 2} Identify D numbers of these three evidence bodies respectively. We first choose a scenario from Fig. \ref{bpa_figure}, in which the following bodies of evidence are observed:
\begin{description}
    \item[(1)] Pipe breakage rate is `10 breaks/100 km/year' (Fig. \ref{bpa_figure}(a)).
    $$D_1(P)=0 \quad D_1(P, NP)=0 \quad D_1(NP)=1$$
    \item[(2)] The possibility of pressure drop to `0 psi' at the respective node (Fig. \ref{bpa_figure}(b)).
    $$D_2(P)=0 \quad D_2(P, NP)=1 \quad D_2(NP)=0, \quad and$$
    \item[(3)] A leaky sewer is located at a distance of `3' m (Fig. \ref{bpa_figure}(c)).
    $$D_3(P)=1 \quad D_3(P, NP)=0 \quad D_3(NP)=0.$$
\end{description}

\textbf{Step 3} Following steps in section \ref{exclusive section} to calculate the exclusive coefficient of these evidence bodies respectively. Results are shown as follows.
      $$\varepsilon_1=0.1195, \quad \varepsilon_2=0.1057, \quad \varepsilon_1=0.131$$

\textbf{Step 4} Discount D numbers above according to Eq. \ref{discount}. Table \ref{scenario 1} also shows the discounted BPA.
\begin{table}[htbp]
\centering
\def\~{\hphantom{0}}
\caption{Assessment of the risk of contaminant intrusion.}
\label{scenario 1}
\begin{tabular*}{\columnwidth}{@{\extracolsep{\fill}}@{~~}cccccc@{~~}}
\hline
Evidence & value  & $\varepsilon\quad$  & $D_i{(\{P\})}_\varepsilon$ & $D_i{(\{P, NP\})}_\varepsilon$ & $D_i{(\{NP\})}_\varepsilon$\\
\hline
 Pathway,$e_1$  & 10 & 0.1195 & 0	& 0.869 & 0.131\\(\# bks/100 km/year)\\
 Pressure, $e_2$(psi) & 0 & 0.1057 & 0 & 0 & 1\\
 Contaminant source,  & 3 &0.131 & 0.869 & 0 & 0.131\\ $e_3$(m)\\
\hline
\end{tabular*}
\end{table}

\textbf{Step 5} Use the combination rule of D numbers (Eq. \ref{conbination}) to obtain integrated D numbers of risk of contaminant intrusion at a given location in a distribution network. For example, the integrated D numbers of this scenario for risk assessment is:
      $$D(\{P\})=0.44,\quad D(\{NP\})=0.49,\quad D(\{P,NP\})=0.07 $$

To illustrate the efficiency of the proposed method, we choose five additional scenarios as \cite{sadiq2006estimating} do. The results obtained through both methods are illustrated in Table \ref{tab:results}.
\begin{table}[htbp]
\centering
\small
\begin{threeparttable}
\caption{\label{tab:results}Some hypothetical scenarios to demonstrate the comparison of two methods.}
\begin{tabular}{lccccc}
\toprule
Scenario & $e_1$  & $e_2$ & $e_3$ & Method & Risk\\&&&&&$(\{P\},\{P,NP\},\{NP\})$\\
\midrule
1 & 10 & 0 & 3 & Sadiq \emph{et al}. (2006) & $(0.31, 0.21, 0.48)$\\&&&& Proposed &(0.44, 0.07, 0.49)\\
\midrule
2 & 10 & 0 & 20 & Sadiq \emph{et al}. (2006) & $(0, 0.3, 0.7)$\\&&&& Proposed &(0, 0.1195, 0.8805)\\
\midrule
3 & 10 & 50 & 20 & Sadiq \emph{et al}. (2006) & $(0, 0.03, 0.97)$\\&&&& Proposed &(0, 0.013, 0.987)\\
\midrule
4 & 30 & 50 & 20 & Sadiq \emph{et al}. (2006) & $(0, 0.1, 0.9)$\\&&&& Proposed &(0,0.11,0.89)\\
\midrule
5 & 30 & 50 & 3 & Sadiq \emph{et al}. (2006) & $(0.13, 0.09, 0.78)$\\&&&& Proposed &(0.41,0.06,0.53)\\
\midrule
6 & 30 & -20 & 3 & S & $(0.96, 0.04, 0)$\\&&&& Proposed &(0.98,0.02,0)\\
\bottomrule
\end{tabular}
\begin{tablenotes}
\item[] e1: Breaks (\# bks/100 km/year)\\
e2: Pressure (psi)\\
e3: Separation distance (m)
\end{tablenotes}
\end{threeparttable}
\end{table}
The proposed model make the assessment based on D numbers discounted by the exclusive coefficient, which reflects the exclusive degree of these propositions. While the method in \cite{sadiq2006estimating} use the Dempster-Shafer evidence theory based on the mutually exclusive hypothesis between the propositions, which is impractical in the real-word applications. From the results, we can see that, the incompleteness of the result on the bias of exclusive coefficient is much smaller than that of \cite{sadiq2006estimating}. The proposed method not only allow the propositions of application to be non-exclusive, but also give a more effective assessment.

%

\section{Conclusion}
One of the assumptions to apply the Dempster-Shafer evidence evidence theory is that all the elements in the frame of discernment should be mutually exclusive. However, it is difficult to meet the requirement in the real-world applications. In this paper, a new mathematic tool to model uncertain information, called as D numbers, is used to model and combine the domain experts' opinions under the condition that the linguistic constants are not exclusive with each other. An exclusive coefficient is proposed to discount the D numbers. After the discounted D numbers are obtained, the domain experts' opinion can be fused based on our proposed combination rule of D numbers. The application to estimate the risk of the contamination intrusion of the water distribution network illustrates the efficiency of our proposed D numbers method.

\section*{Acknowledgements}
The work is partially supported by National Natural Science Foundation of China (Grant No. 61174022), R\&D Program of China (2012BAH07B01), National High Technology Research and Development Program of China (863 Program) (Grant Nos. 2013AA013801 and  2012AA041101), Chongqing Natural Science Foundation (for Distinguished Young Scholars) (Grant No. CSCT,2010BA2003), Doctor Funding of Southwest University (Grant No. SWU110021), China State Key Laboratory of Virtual Reality Technology and Systems.

\section*{References}

\bibliography{mybibfile}

\begin{thebibliography}{10}
\expandafter\ifx\csname url\endcsname\relax
  \def\url#1{\texttt{#1}}\fi
\expandafter\ifx\csname urlprefix\endcsname\relax\def\urlprefix{URL }\fi
\expandafter\ifx\csname href\endcsname\relax
  \def\href#1#2{#2} \def\path#1{#1}\fi

\bibitem{sargaonkar2013model}
A.~Sargaonkar, S.~Kamble, R.~Rao, Model study for rehabilitation planning of
  water supply network, Computers, Environment and Urban Systems 39 (2013)
  172--181.

\bibitem{nyende2013application}
S.~Nyende-Byakika, G.~Ngirane-Katashaya, J.~M. Ndambuki, Application of
  hydraulic modelling to control intrusion into potable water pipelines, Urban
  Water Journal 10~(3) (2013) 216--219.

\bibitem{vairavamoorthy2007modelling}
K.~Vairavamoorthy, J.~Yan, S.~Gorantiwar, Modelling the risk of contaminant
  intrusion in water mains, Proceedings of the ICE-Water Management 160~(2)
  (2007) 123--132.

\bibitem{preis2008multiobjective}
A.~Preis, A.~Ostfeld, Multiobjective contaminant response modeling for water
  distribution systems security, Journal of Hydroinformatics 10~(4) (2008)
  267--274.

\bibitem{tamminen2008water}
S.~Tamminen, H.~Ramos, D.~Covas, Water supply system performance for different
  pipe materials part {I}: water quality analysis, Water resources management
  22~(11) (2008) 1579--1607.

\bibitem{perelman2009extreme}
L.~Perelman, A.~Ostfeld, Extreme impact contamination events sampling for water
  distribution systems security, Journal of Water Resources Planning and
  Management 136~(1) (2009) 80--87.

\bibitem{islam2013evaluating}
M.~S. Islam, R.~Sadiq, M.~J. Rodriguez, H.~Najjaran, A.~Francisque, M.~Hoorfar,
  Evaluating water quality failure potential in water distribution systems: a
  fuzzy-topsis-owa-based methodology, Water resources management 27~(7) (2013)
  2195--2216.

\bibitem{el2004fuzzy}
I.~El-Baroudy, S.~P. Simonovic, Fuzzy criteria for the evaluation of water
  resource systems performance, Water resources research 40~(10).

\bibitem{walski2003advanced}
T.~M. Walski, D.~V. Chase, D.~A. Savic, W.~M. Grayman, S.~Beckwith, E.~Koelle,
  et~al., Advanced water distribution modeling and management, Haestad press,
  2003.

\bibitem{islam2013evaluation}
N.~Islam, R.~Sadiq, M.~J. Rodriguez, A.~Francisque, Evaluation of source water
  protection strategies: A fuzzy-based model, Journal of environmental
  management 121 (2013) 191--201.

\bibitem{xin2012hazard}
K.~Xin, T.~Tao, Y.~Wang, S.~Liu, Hazard and vulnerability evaluation of water
  distribution system in cases of contamination intrusion accidents, Frontiers
  of Environmental Science \& Engineering 6~(6) (2012) 839--848.

\bibitem{khanal2006distribution}
N.~Khanal, S.~G. Buchberger, S.~A. McKenna, Distribution system contamination
  events: exposure, influence, and sensitivity, Journal of water resources
  planning and management 132~(4) (2006) 283--292.

\bibitem{lindley2001framework}
T.~R. Lindley, A framework to protect water distribution systems against
  potential intrusions, Ph.D. thesis, University of Cincinnati (2001).

\bibitem{lindley2002assessing}
T.~R. Lindley, S.~G. Buchberger, et~al., Assessing intrusion susceptibility in
  distribution systems, Journal-American Water Works Association 94~(6) (2002)
  66--79.

\bibitem{rasekh2014drinking}
A.~Rasekh, K.~Brumbelow, Drinking water distribution systems contamination
  management to reduce public health impacts and system service interruptions,
  Environmental Modelling \& Software 51 (2014) 12--25.

\bibitem{shen2011false}
H.~Shen, E.~McBean, False negative/positive issues in contaminant source
  identification for water-distribution systems, Journal of Water Resources
  Planning and Management 138~(3) (2011) 230--236.

\bibitem{preis2007contamination}
A.~Preis, A.~Ostfeld, A contamination source identification model for water
  distribution system security, Engineering optimization 39~(8) (2007)
  941--947.

\bibitem{sadiq2008predicting}
R.~Sadiq, E.~Saint-Martin, Y.~Kleiner, Predicting risk of water quality
  failures in distribution networks under uncertainties using fault-tree
  analysis, Urban Water Journal 5~(4) (2008) 287--304.

\bibitem{zadeh1984review}
L.~A. Zadeh, Review of a mathematical theory of evidence, AI magazine 5~(3)
  (1984) 81.

\bibitem{sadiq2006estimating}
R.~Sadiq, Y.~Kleiner, B.~Rajani, Estimating risk of contaminant intrusion in
  water distribution networks using dempster--shafer theory of evidence, Civil
  Engineering and Environmental Systems 23~(3) (2006) 129--141.

\bibitem{jenelius2010critical}
E.~Jenelius, J.~Westin, {\AA}.~J. Holmgren, Critical infrastructure protection
  under imperfect attacker perception, International Journal of Critical
  Infrastructure Protection 3~(1) (2010) 16--26.

\bibitem{zadeh1965fuzzy}
L.~A. Zadeh, Fuzzy sets, Information and control 8~(3) (1965) 338--353.

\bibitem{deng2012fuzzy}
Y.~Deng, Y.~Chen, Y.~Zhang, S.~Mahadevan, Fuzzy dijkstra algorithm for shortest
  path problem under uncertain environment, Applied Soft Computing 12~(3)
  (2012) 1231--1237.

\bibitem{zhang2013ifsjsp}
X.~Zhang, Y.~Deng, F.~T. Chan, P.~Xu, S.~Mahadevan, Y.~Hu, {IFSJSP}: A novel
  methodology for the job-shop scheduling problem based on intuitionistic fuzzy
  sets, International Journal of Production Research 51~(17) (2013) 5100--5119.

\bibitem{aghaarabi2014compara}
E.~Aghaarabi, F.~Aminravan, R.~Sadiq, M.~Hoorfar, M.~Rodriguez, H.~Najjaran,
  Comparative study of fuzzy evidential reasoning and fuzzy rule-based
  approaches: an illustration for water quality assessment in distribution
  networks, Stochastic Environmental Research and Risk Assessment 28~(3) (2014)
  655--679.

\bibitem{setola2009critical}
R.~Setola, S.~De~Porcellinis, M.~Sforna, Critical infrastructure dependency
  assessment using the input--output inoperability model, International Journal
  of Critical Infrastructure Protection 2~(4) (2009) 170--178.

\bibitem{oliva2011fuzzy}
G.~Oliva, S.~Panzieri, R.~Setola, Fuzzy dynamic input--output inoperability
  model, International Journal of Critical Infrastructure Protection 4~(3)
  (2011) 165--175.

\bibitem{pawlak2007rudiments}
Z.~Pawlak, A.~Skowron, Rudiments of rough sets, Information sciences 177~(1)
  (2007) 3--27.

\bibitem{pawlak2007rough}
Z.~Pawlak, A.~Skowron, Rough sets: some extensions, Information sciences
  177~(1) (2007) 28--40.

\bibitem{dempster1967upper}
A.~P. Dempster, Upper and lower probabilities induced by a multivalued mapping,
  The annals of mathematical statistics (1967) 325--339.

\bibitem{shafer1976mathematical}
G.~Shafer, A mathematical theory of evidence, Vol.~1, Princeton university
  press Princeton, 1976.

\bibitem{wei2013identifying}
D.~Wei, X.~Deng, X.~Zhang, Y.~Deng, S.~Mahadevan, Identifying influential nodes
  in weighted networks based on evidence theory, Physica A: Statistical
  Mechanics and its Applications 392~(10) (2013) 2564--2575.

\bibitem{deng2014environmental}
X.~Deng, Y.~Hu, Y.~Deng, S.~Mahadevan, Environmental impact assessment based on
  {D} numbers, Expert Systems with Applications 41~(2) (2014) 635--643.

\bibitem{huang2013new}
S.~Huang, X.~Su, Y.~Hu, S.~Mahadevan, Y.~Deng, A new decision-making method by
  incomplete preferences based on evidence distance, Knowledge-Based
  Systems~(56) (2014) 264--272.

\bibitem{li2007hierarchical}
H.~Li, Hierarchical risk assessment of water supply systems, Ph.D. thesis
  (2007).

\bibitem{weickgenannt2010risk}
M.~Weickgenannt, Z.~Kapelan, M.~Blokker, D.~A. Savic, Risk-based sensor
  placement for contaminant detection in water distribution systems, Journal of
  Water Resources Planning and Management 136~(6) (2010) 629--636.

\bibitem{sadiq2004aggregative}
R.~Sadiq, Y.~Kleiner, B.~Rajani, Aggregative risk analysis for water quality
  failure in distribution networks., Journal of Water Supply: Research \&
  Technology-AQUA 53~(4) (2004) 241--261.

\bibitem{deng2011modeling}
Y.~Deng, W.~Jiang, R.~Sadiq, Modeling contaminant intrusion in water
  distribution networks: A new similarity-based dst method, Expert Systems with
  Applications 38~(1) (2011) 571--578.

\bibitem{deng2012d}
Y.~Deng, D numbers: Theory and applications, Journal of Information and
  Computational Science 9~(9) (2012) 2421--2428.

\bibitem{Deng2014DAHPSupplier}
X.~Deng, Y.~Hu, Y.~Deng, S.~Mahadevan, Supplier selection using {AHP}
  methodology extended by {D} numbers, Expert Systems with Applications 41~(1)
  (2014) 156--167.

\bibitem{Deng2014BridgeDNs}
X.~Deng, Y.~Hu, Y.~Deng, Bridge condition assessment using {D} numbers, The
  Scientific World Journal 2014 (2014) Article ID 358057, 11 pages.
\newblock \href {http://dx.doi.org/10.1155/2014/358057}
  {\path{doi:10.1155/2014/358057}}.

\bibitem{tao2012identification}
T.~Tao, Y.-j. Lu, X.~Fu, K.-l. Xin, Identification of sources of pollution and
  contamination in water distribution networks based on pattern recognition,
  Journal of Zhejiang University SCIENCE A 13~(7) (2012) 559--570.

\bibitem{deng2011risk}
Y.~Deng, R.~Sadiq, W.~Jiang, S.~Tesfamariam, Risk analysis in a linguistic
  environment: a fuzzy evidential reasoning-based approach, Expert Systems with
  Applications 38~(12) (2011) 15438--15446.

\end{thebibliography}

\end{document}